\documentclass{bmvc2k}

\title{Localization Guided Learning for \\Pedestrian Attribute Recognition}

\addauthor{Pengze Liu}{pengzeliu2-c@my.cityu.edu.hk}{1}
\addauthor{Xihui Liu}{xihui-liu@link.cuhk.edu.hk}{2}
\addauthor{Junjie Yan}{yanjunjie@sensetime.com}{3}
\addauthor{Jing Shao}{shaojing@sensetime.com}{3}

\addinstitution{
 Department of Computer Science\\
 City University of Hong Kong\\
 Hong Kong, China
}
\addinstitution{
 Department of Electronic Engineering\\
 The Chinese University of Hong Kong\\
 Hong Kong, China
}
\addinstitution{
 SenseTime Research\\
 Beijing, China
}

\runninghead{Liu \etal}{Localization Guided Learning for 
Pedestrian Attribute}

\def\etal{\emph{et al}\bmvaOneDot}
\usepackage{amssymb}
\newcommand{\R}{\mathbb{R}}
\begin{document}

\maketitle

\begin{abstract}
Pedestrian attribute recognition has attracted many attentions due to 
its wide applications in scene understanding and person analysis from 
surveillance videos. Existing methods try to use additional pose, part or 
viewpoint information to complement the global feature representation for 
attribute classification. However, these methods face difficulties in localizing 
the areas corresponding to different attributes. To address this problem, we 
propose a novel Localization Guided Network which assigns attribute-specific 
weights to local features based on the affinity between proposals 
pre-extracted proposals and attribute locations. The advantage of our 
model is that our local features are learned automatically for each attribute 
and emphasized by the interaction with global features. We demonstrate 
the effectiveness of our Localization Guided Network on two pedestrian 
attribute benchmarks (PA-100K and RAP). Our result surpasses the 
previous state-of-the-art in all five metrics on both datasets.
\end{abstract}

\section{Introduction}
\label{sec:intro}
Pedestrian attribute recognition is a high-demanding problem due 
to its wide applications in person re-identification, person 
retrieval, and social behavioral analysis. Boosted by the 
increasing demand, three pedestrian attribute benchmarks 
~\cite{deng2014pedestrian,li2016richly,liu2017hydraplus} are 
released to facilitate the revolution of pedestrian attribute recognition 
approaches. Inspired by the increasing amount of annotated data, a sequence 
of deep learning based methods~\cite{sudowe2015person,li2015multi,
liu2017hydraplus,wang2017attribute, sarfraz2017deep} are proposed with 
remarkable advantages over previous SVM-based approaches. Most 
recently-proposed methods formulate pedestrian attribute recognition 
as a multi-label classification problem, and address it by optimizing 
a parameter-sharing convolutional network to exploit semantic relations 
among attributes in an adaptive manner, which achieves a huge success. 

The performances of these deep learning methods are constrained by 
several factors. Firstly, the resolution of pedestrian images is not 
admirable. Specifically, the lowest resolutions of images are $36\times92$ 
for RAP and $50\times100$ for PA-100K. Moreover, the scales of 
attributes vary from the whole image (e.g. "fat", "thin") to small 
regions (e.g. "glasses", "shoes"). It could be especially challenging 
to classify some small attributes due to the information loss caused by 
down-sampling. Secondly, due to the divergence of the angles and 
distances between the pedestrians and the surveillance cameras, 
as well as the imperfection of pedestrian detection algorithms, 
the pedestrian images have great variation in shapes and viewpoints.
The recognition difficulty increases due to the variation 
in viewpoints, as introduced by Sarfraz \etal~\cite{sarfraz2017deep}. 
\begin{figure}[]
\begin{center}
\includegraphics[width=12.56cm]{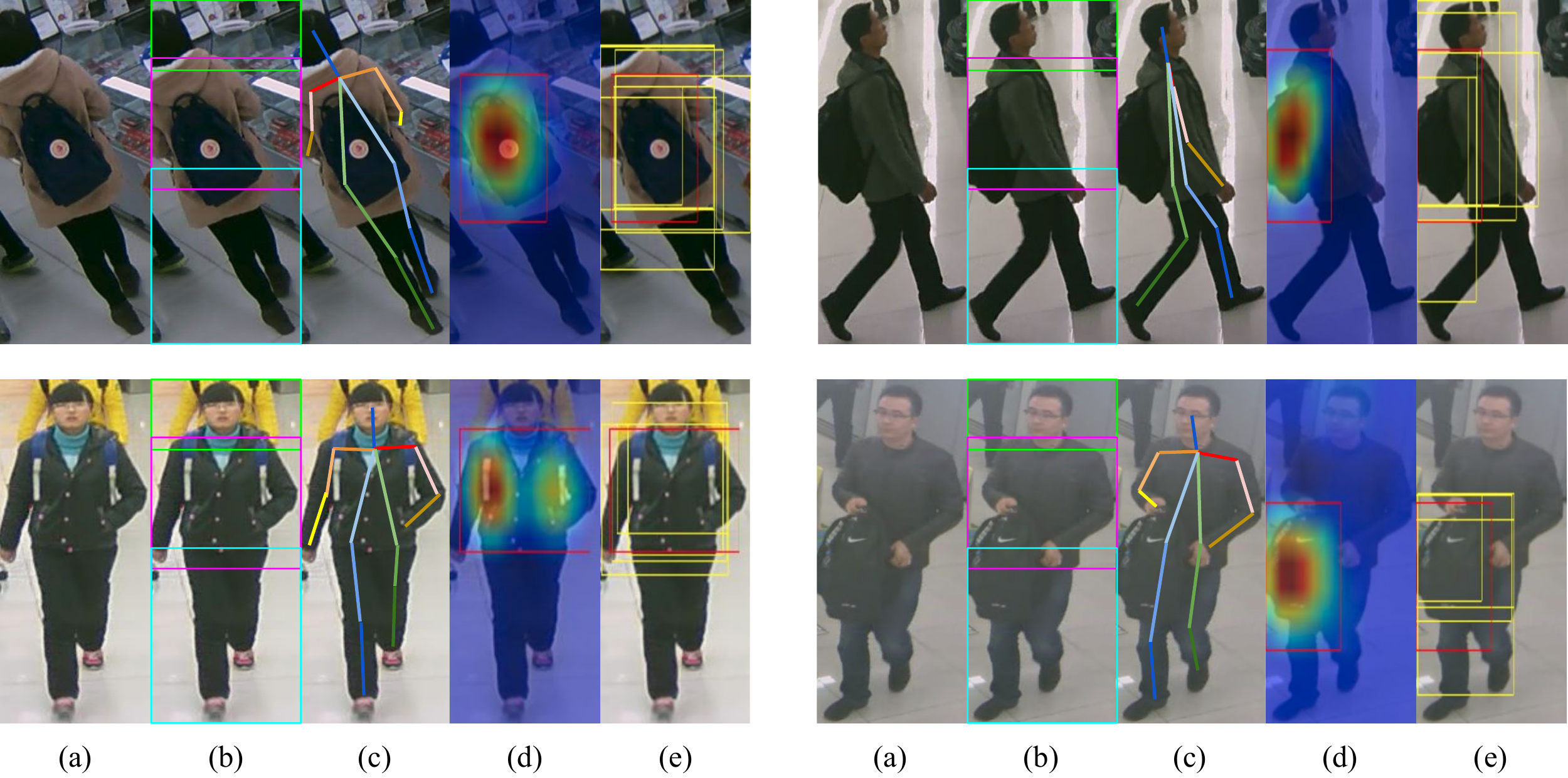}
\end{center}
\caption{Pedestrian attribute recognition with local features relies on 
the accurate localization of attributes. Take backpack as an example, we 
present schematic diagrams of different methods, i.e. part-based method 
(b), keypoint-based method (c), activation map-based method (d) and our 
localization guided method (e, represented by top-5 proposals). Result 
suggests that the object is better localized in our method regardless of 
the variance in the viewpoints and the relative positions of objects. 
Meanwhile, our method is in favor of the object and its surrounding areas, 
compared with only the object in the class activation map.}
\label{fig:fig1}
\end{figure}
For example, Figure 1(a) shows some examples of backpack recognition. 
The original images in column (a) show images from three viewpoints, and 
the backpack appears in hand in the last image. The large variance of
relative positions between the pedestrians and these backpacks suggests 
that it is hard to localize objects only by its relative position 
regardless of the input image. 

One intuitive solution to address the problems mentioned above is 
to apply local feature mining in order to fully explore the 
information from specific regions. Most existing CNN-based methods use
either human part or pose information to extract local features for 
multiple pedestrian recognition tasks. For part information, one typical 
solution as introduced in ~\cite{fabbri2017generative,li2017learning} is 
to split the image vertically into upper, middle and lower part, and 
feed each part into a distinct network for feature extraction. 
The features are merged by concatenating with a global feature vector 
learned from the whole image for end-to-end optimization. One problem with 
these methods is that the splitting of the body part is defined by a fixed, 
human-specific strategy that can be imprecise. It may lead to truncated 
parts for some images. To better exploit the human part in the images, pose 
information is introduced in~\cite{park2016attribute,zhao2017spindle} that 
utilize a landmark model for body part localization. While due to the 
imperfection of human detector quality and pose variations, the missing body 
part in the image could confuse such models. Another solution is to design 
a network based on attention models. Liu \etal~\cite{liu2017hydraplus} 
introduced a soft attention model that extract and concatenate 
multi-level global and local feature patterns for attribute classification. 

The common constraint of the methods mentioned above is that they do not 
specifically locate attributes in the images, but use human-defined 
local feature extraction strategies. These methods surpassed ordinary 
results that only extracts global information, indicating that the exploited 
local information can boost the recognition of attributes to some extent. 
To overcome the drawbacks of previous methods and fulfill our 
expectations, we propose a novel network architecture, referred 
as Localization Guided Network (LG-Net) as shown in Figure 2 to 
fully exploit the attribute-specific local features for attribute 
predictions. Specifically, we modify the class activation 
map~\cite{zhou2016learning} to dynamically generate localization 
result for the accurate localization of attributes. The localization 
of a certain attribute is further projected to the local feature maps 
to achieve attribute-wise local feature extraction. 
The extracted local features are merged with global features in 
an adaptive manner. 
Compared with previous methods, our architecture could 
accurately address the locations of certain attributes, especially 
object-type attributes, and match the surrounding local features for 
location-specific attribute recognition. We evaluate our result on the 
two largest pedestrian datasets by five matrices, showing that 
our goal is achieved with a significant improvement in 
attribute prediction result.

\section{Related Work}
\textbf{Pedestrian Attribute Recognition.} 
Pedestrian attribute recognition is a multi-label classification 
problem that has been widely applied to the person retrieval
and person re-identification. 
Earlier works address this problem with some traditional methods. 
Gray \etal~\cite{gray2008viewpoint} apply AdaBoost algorithm to address 
this problem. Prosser \etal~\cite{prosser2010person}, Layne \etal 
~\cite{layne2012person,layne2014attributes}, Deng \etal 
~\cite{deng2014pedestrian} and Li \etal~\cite{li2016richly} solve this 
problem by optimizing Support Vector Machines (SVMs) to classify single attributes.
The recognition accuracy is significantly 
improved with the recent development of Deep Learning. Sudowe 
\etal~\cite{sudowe2015person} first introduced a convolutional neural 
network with parameter sharing in most layers to adaptively explore 
the semantic relations among attributes. The network is trained 
end-to-end with independent loss layers for each attribute. 

Some recent methods like Yu \etal~\cite{yu2015multi} and Liu \etal 
~\cite{liu2017hydraplus} capture multi-level features from the network to 
exploit global and local contents. Li \etal~\cite{li2015multi} design a 
weighted loss to address the problem of imbalanced positive and negative 
samples. Sarfraz \etal~\cite{sarfraz2017deep} introduce a model with view 
guidance to make view-specific attribute predictions so as to overcome  
the variance of patterns in different angles. Wang \etal 
~\cite{wang2017attribute} proposed a recurrent neural network to explore 
intra-person, inter-person, and inter-attribute context information to better 
exploit semantic relations among attributes. 

Other methods further improve the model design with the pose 
or part information. Zhang \etal~\cite{zhang2014panda} 
explore poselets for part localization and attribute classification. 
Fabbri \etal~\cite{fabbri2017generative}, Li \etal~\cite{li2017learning} 
and Zhu \etal~\cite{zhu2017multi} split input image into fixed parts by 
some handcraft standard to extract features from certain part. Park \etal 
~\cite{park2016attribute} and Zhao \etal~\cite{zhao2017spindle} explore 
key-points of the person to dynamically capture the body pose for part split 
and local feature extraction. Yang \etal~\cite {yang2016attribute} use 
key-points to generate adaptive parts. 
These methods utilize different information to extract local features. 
But the attribute-specific localization information is missing in these 
solutions, which makes these methods less robust to the variance of the 
relative position of the objects of interest. 

\textbf{Weakly Supervised Localization.} 
Weakly supervised localization using only image-level supervision 
is an important problem to reduce the cost of annotations. Some 
methods~\cite{kumar2010self,deselaers2012weakly,cinbis2017weakly,
song2014learning,bilen2014weakly,oquab2015object,wei2016hcp} formulate 
this problem as multiple instance learning (MIL). Oquab \etal 
~\cite{oquab2014weakly} utilize mid-level 
representations that solve object localization with CNN outputs. 
Oquab \etal~\cite{oquab2015object} further develop a CNN that uses global 
max pooling on top of features to localize a point on objects. Similarly, 
Zhou \etal~\cite{zhou2016learning} introduce global average pooling 
to obtain a class activation map. Bounding boxes covering the highlight 
areas are adopted as the localization result. 
Another solution is to cluster the similar patterns among 
regions of interests. Song \etal~\cite{song2014learning} borrow the idea 
from nearest neighbor to select a set of adjacent windows with graph-based 
algorithms. Bilen \etal~\cite{bilen2016weakly} propose a network containing 
two parallel branches with different normalization to perform 
classification and weakly supervised detection respectively. Bency \etal 
~\cite{bency2016weakly} compare different candidate localization to confirm 
boxes as localization results progressively. Bazzani \etal 
~\cite{bazzani2016self} 
and Singh \etal~\cite{singh2017hide} enhance the performance by randomly 
hide some spatial regions for the classification network to capture more 
relevant patterns. Li \etal~\cite{li2018tell} extend this idea 
by filtering the highlight areas instead of random regions 
to perform accurate weakly-supervised segmentation.

\section{Localization Guided Network Architecture}
In order to extract and utilize the local features around the locations 
of attributes, we design a novel architecture for localization guided 
feature extraction and attribute classification, as shown in 
Figure 2. The network is composed of two separate branches, i.e. 
global branch and local branch. The global branch takes image-level 
input to generate localization for all attributes, and the local branch 
utilizes localization to predict the attributes. Both global and local branches 
are adapted from the \textit{Inception-v2}~\cite{ioffe2015batch} 
architecture for its great extensibility in multi-scale feature 
extraction.  

\subsection{Global Feature Extraction}\label{subsec:global_feature_extraction}
The global branch takes the whole image as input and extracts class activation 
maps~\cite{zhou2016learning}, which can provide guidance on localizing attributes.

\textbf{Class Activation Box}. Class activation box is obtained from the class 
activation map, acting as the locations of attributes in our method, as shown 
in Figure 2. Specifically, the original CAM method takes 
the weighted sum of the output feature maps as the class activation map of a 
class, where the weights are corresponding to the weights between 
classes and the output units of the global average pooling (GAP) layer. 
In our method, we implement this in an equivalent way, to 
enable calculating class activation maps in a forward pass during training and 
testing. We extract the weight from a pretrained classification model and use the 
extracted weight to initialize the class activation generator, implemented as a 
fixed $1 \times 1$ convolution layer on top of the 
global branch. The convolution layer dynamically produces $c$ class activation maps 
for each image, where $c$ is the number of attributes in our dataset. In order to keep 
the activation map unchanged, we fix the parameter updating in the global branch. 
After the class activation maps are obtained, an activation box is captured for each 
attribute by cropping the high-response areas of the corresponding activation map. 
The activation boxes act as the localization result in our method.

\subsection{Local Feature Extraction}\label{subsec:loca_feature_extraction}
The local branch serves the purpose of extracting local features 
from the input image. In our method, we use EdgeBoxes~\cite{zitnick2014edge} 
to generate region proposals. Specifically, for each input image \textbf{x}, 
EdgeBoxes output a list of candidate regions $R = (R_1,...,R_n)$ 
where $R_i$ is a 5-dimensional array, 
containing a bounding box and a score for each box. We extract 
a fixed number of proposals for each image. The local branch 
extracts local features by an ROI pooling layer between 
\textit{inception-5a} and \textit{inception-5b}, such that a sufficient 
amount of information is provided for proposal processing while the 
overall computational cost is reasonable. 

The extracted local features are fed into \textit{inception-5b} 
and a following global average pooling for further processing. The 
outputs of the network are $n$ feature vectors for each image where 
$n$ is the number of proposals. All proposals are even in the feature 
extraction part. The $n$ proposal regions could cover most possible 
locations of attributes, local features are extracted into separate 
vectors for future utilization.

\begin{figure}[]
\begin{center}
\includegraphics[width=12.56cm]{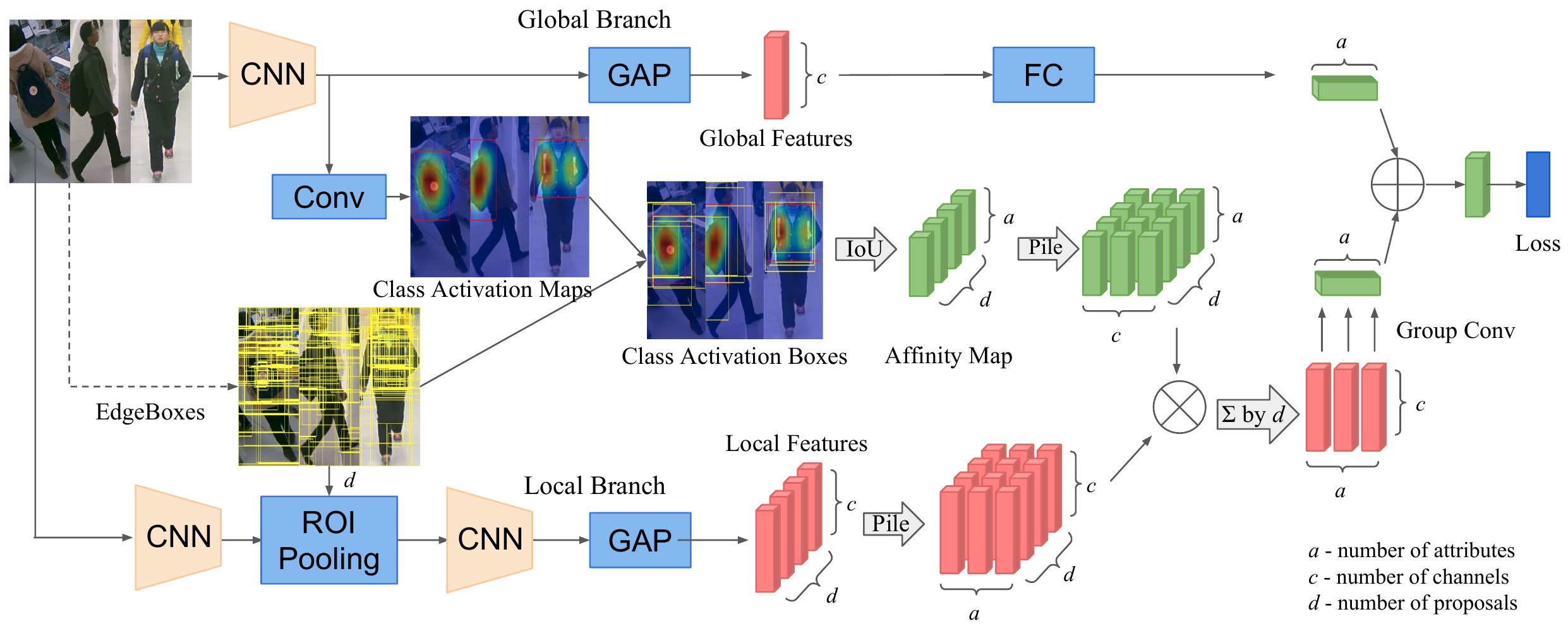}
\end{center}
\caption{Our Localization Guided Network is composed 
of two branches, i.e. global branch and local branch. The global branch is a 
parameter-fixed branch, taking the whole image as input to generate attribute 
locations based on the class activation map. The convolution layer is 
fixed and initialized by the weights of the fully-connected layer in pretrained model. 
Localization results are used for calculating the weights to the local features for 
attribute-wise local feature selections. Weighted local features are projected to a 
single vector by attribute. The element-wise summation of global and local feature 
vectors are used for attribute predictions.}
\label{fig:fig2}
\end{figure}

\subsection{Localization Guidance Module}

In Section~\ref{subsec:loca_feature_extraction}, we have extracted local features of all proposals. However, it is natural to consider that the proposals should not contribute equally, and different attributes should focus on different local features. For example, for the attribute 'wear glasses', the local features around head should contribute more, while for the attribute 'color of shoes', usually it relates to local features at the bottom. So the intuition is that we should decide how much each local feature should contribute to the classification of a certain attribute. Fortunately, in Section \ref{subsec:global_feature_extraction}, we have obtained the class activation map for each attribute, which could serve as a guide to determine the importance of the local features to different attributes.

So as to generate attribute-specific feature vectors for attribute recognition, 
localization guidance is expected beyond local feature extraction. We 
design a Localization Guidance Module that takes input information from 
both global and local branches to map local feature vectors with attributes 
by the spatial affinity between the proposals and the class activation boxes. 
In the following part, we will introduce the localization of attributes, the 
affinity calculation between ROIs and locations of attributes, together with 
the fusion strategy so as to generate the image-level attribute-specific information 
with local feature maps. 

\textbf{CAM-ROI Affinity Map}. 
Different proposals are of variable importance to different attributes. 
To address the important regions for each attribute, class activation 
boxes generated by CAMs provide the information about the importance 
of proposals. In our method, the affinity between class activation boxes 
and proposals are calculated as the IoU (Intersection over Union) 
between the two bounding boxes as shown in Figure 2, 
due to the need of eliminating the influence 
of proposals with a little intersection with the object locations and punish 
the proposals that are too large to focus on a specific region of the 
attribute. 

Formally, denoting $c$ the number of classes in 
a dataset and $d = |R|$ the number of proposals, for each 
given image, two shortlists of bounding boxes are associated with 
length $c$ generated from CAM~\cite{zhou2016learning} and $d$ generated 
by EdgeBoxes~\cite{zitnick2014edge} respectively. In order to capture 
the proposal boxes with high correlations with attribute 
localization, a CAM-ROI Affinity Map is expected. We formulate 
the Affinity as follows:
$$
A_{i,j} = \frac{C_i \cap D_j}{C_i \cup D_j}
$$
where $C_i$ is the i-th element of the CAM boxes and $D_j$ is the 
j-th element of the ROI regions. The affinity between two 
proposals is calculated by the Intersection over Union (IoU).

An affinity map in $c \times d$ is calculated and linearly normalized 
to weight the local feature vectors for further predictions. There is 
no need for the affinity function to be differentiable since the class 
activation maps are generated from a fixed branch. 

\textbf{Localization Guidance.}
The shapes of weighted local feature tensors are determined by 
the numbers of proposals and the number of attributes in consideration. 
While the shape of the global feature is dependent on the number of 
convolution kernels in the last layer. So as to take the element-wise 
summation from both global and local branch, a fusion module is designed 
to combine the output of two branches, while keeping the correlations 
between the proposals and the attributes consistent.

The Localization Guided Network is composed of a convolutional neural 
network with ROI pooling, followed by a localization guiding block. 
The localization guiding block is introduced as follows. 
For each sample, the input is an image, denoted as $I$, 
and its associated proposals $R$. The global stream 
would produce an affinity map: $$A = f_g(I;\theta_g), A \in 
\R^{d \times c \times 1}$$ Meanwhile, the feature vector
obtained from the local stream after global pooling is reshaped 
into: $$X = f_l(I, R, \theta_l), X \in \R^{d \times 1 \times k}$$
Where $f_g$, $f_l$ are the two global and local streams of the 
LG-Net and $\theta$ the corresponding parameters. So as to apply 
element-wise weighting, $A$ and $X$ are both tiled into $\R^{
d \times c \times k}$ and produced as the localization 
guided local feature matrices $\hat{X} \in \R^{d \times c \times k}$.

The local features need to be merged for image-level recognition. In our 
method, a non-weighted summation of all proposals is used as the image-level 
local feature tensor considering the scale. The image-level feature tensor 
preserves the attribute-wise information in the format of a 1024 dimensional 
feature vector for each attribute of an image. To obtain a $c$ dimensional 
local feature vector for each image, each 1024 dimensional feature vector is 
mapped to a single value by an adaptively weighted sum. The local features are 
then mapped to an attribute-wise local feature vector $\hat{Y}_l \in \R^{c}$.

For the fusion of global and local features, we apply element-wise sum to 
the two mentioned feature matrices:
$$\hat{Y} = \hat{Y}_l \bigoplus \hat{Y}_g$$
The summation act as the classification score for each attribute supervised 
by a weighted sigmoid cross entropy loss, which takes image-level ground 
truth to update parameters by weighted gradient back-propagation. The gradient 
values of the positive samples are multiplied by a weighting factor $w_c$ to 
overcome the data imbalance as introduced in~\cite{li2015multi}.

\subsection{Implementation Details}
In this section, we use $a$ to denote the number of attributes of interest, and 
$c$ to denote the number of feature maps in the last layer of the fundamental 
convolutional network to maintain consistency with Figure 2. The LG-Net is 
trained in two stages. In the first training stage, an Inception-v2 classification 
network is trained from the pedestrian attribute benchmarks, initialized by an 
ImageNet pretrained model. The weight outcome in the first stage is denoted 
as $\theta_\alpha$. The weight of the fully-connected layer in $\theta_\alpha$ 
is extracted as $\theta_\alpha(FC) \in \R^{a \times c}$. 

In the second training stage, $\theta_\alpha$ is acting as the initialization of the 
global branch of a LG-Net, as shown in Figure 2. 
A convolution layer of size $1 \times 1$, stride 1 with $c$ units 
is initialized by $\theta_\alpha(FC)$ on top of the feature maps of the global 
branch. This layer generates class activation maps for each image in the 
forward pass of the network. The network parts mentioned above in the second 
stage are fixed without parameter updating. This setting is to ensure the accurate 
production of class activation boxes for the ultimate localization result. 

The LG-network is optimized by Stochastic Gradient Descent (SGD)
~\cite{bottou2012stochastic} with an initial learning rate of 0.02 and a weight 
decay of 0.005. The learning rate decays by 0.1 every 20 epochs. 
We train the network for 50 epochs and select the best model based on the 
performance on the validation set. For each image, we select top 100 proposals 
generated by EdgeBoxes~\cite{zitnick2014edge} and post-processed by 
Non-Maximum Suppression. The activated region for an attribute is defined 
as the pixels whose activation value on the activation map is greater than 0.2 
times the maximum activation value in the image for the particular attribute. 

\section{Experiment}
\subsection{Benchmark Data}
RAP~\cite{li2016richly} and PA-100K~\cite{liu2017hydraplus} are the two 
largest public pedestrian attribute datasets. The Richly Annotated 
Pedestrian (RAP) dataset contains 41,585 images collected from indoor 
surveillance cameras. Each image is annotated with 72 attributes,
while only 51 binary attributes with the positive ratio above 1\% are 
selected for evaluation. There are 33,268 images for the training 
set and 8,317 for testing. PA-100K is a recent-proposed large 
pedestrian attribute dataset, with 100,000 images in total 
collected from outdoor surveillance cameras. It is split into 
80,000 images for the training set, and 10,000 for the validation set and 
10,000 for the test set. This dataset is labeled by 26 binary 
attributes. The common features existing in both selected dataset 
is that the images are blurry due to the relatively low 
resolution and the positive ratio of each binary attribute is low.

Following the settings of previous arts, we 
apply both \textit{label-based} evaluation by calculating the 
mean Accuracy (mA) by averaging the accuracy on the positive and 
negative samples, and \textit{sample-based} metrics including 
accuracy, precision, recall, and F1. These metrics are applied to 
our comparison on both datasets.

\subsection{Quantitative Comparison with Prior Arts} 
\begin{table}[]
\centering \scriptsize
\begin{tabular}{|l|ccccc||ccccc|}
\hline
Dataset          & \multicolumn{5}{c||}{RAP} & \multicolumn{5}{c|}{PA-100K}             \\ \hline
Method          & mA    & Accu  & Prec  & Recall & F1    & mA    & Accu  & Prec  & Recall & F1   \\ \hline
ELF+SVM~\cite{prosser2010person}  & 69.94 & 29.29 & 32.84 & 71.18  & 44.95 & -     & -     & -     & -     & -     \\ \cline{1-1} 
CNN+SVM~\cite{li2016richly}  & 72.28 & 31.72 & 35.75 & 71.78  & 47.73 & -     & -     & -     & -     & -     \\ \cline{1-1} 
ACN~\cite{sudowe2015person}  & 69.66 & 62.61 & 80.12 & 72.26  & 75.98 & -     & -     & -     & -     & -     \\ \cline{1-1} 
DeepMar~\cite{li2015multi}  & 73.79 & 62.02 & 74.92 & 76.21  & 75.56 & 72.70 & 70.39 & 82.24 & 80.42 & 81.32 \\ \cline{1-1} 
HP-Net~\cite{liu2017hydraplus}  & 76.12 & 65.39 & 77.33 & 78.79  & 78.05 & 74.21 & 72.19 & 82.97 & 82.09 & 82.53 \\ \cline{1-1} 
JRL~\cite{wang2017attribute}  & 77.81 & -     & 78.11 & 78.98  & 78.58 & -     & -     & -     & -     & -     \\ \cline{1-1} 
VeSPA~\cite{sarfraz2017deep}  & 77.70 & 67.35 & 79.51 & 79.67  & 79.59 & 76.32 & 73.00 & 84.99 & 81.49 & 83.20 \\ \hline
Inception-v2~\cite{ioffe2015batch}  & 75.43 & 65.94 & 79.78 & 77.05  & 78.39 & 72.65 & 71.56 & 84.12 & 80.30 & 82.17 \\ \cline{1-1} 
LG-Net  & \textbf{78.68} & \textbf{68.00} & \textbf{80.36} & \textbf{79.82} & \textbf{80.09} & \textbf{76.96} & \textbf{75.55} & \textbf{86.99} & \textbf{83.17} & \textbf{85.04} \\ \hline
\end{tabular}
\caption{Quantitative Comaprison with State-of-the-Art. Our method surpasses 
previous state-of-the-art by all five metrics on both datasets.}
\end{table}

We demonstrate the effectiveness on both datasets by keeping the 
same settings with previous methods. Table 1 
shows the comparison of our approach with all available public 
datasets. Our proposed method is named in LG-Net, referring to 
the Localization Guided Network. We also list the result of our 
Inception-v2 baseline model.

The result suggests that our LG-Net surpasses all 
previous methods in both label-based and sample-based metrics on 
RAP dataset. Since PA-100K is a newly proposed dataset, it is 
less likely to compare all the previous pedestrian attribute 
methods. Following the evaluation method in~\cite{liu2017hydraplus}, 
we replicate the state-of-the-art method, known as VeSPA on PA-100K, 
and compare our result with previous state-of-the-art and 
previously released results. Our LG-Net significantly surpasses 
VeSPA and earlier methods, suggesting that our method achieves
the state-of-the-art result on PA-100K.

\subsection{Qualitative Evaluation} 
\begin{figure}
\begin{center}
\includegraphics[width=12.56cm]{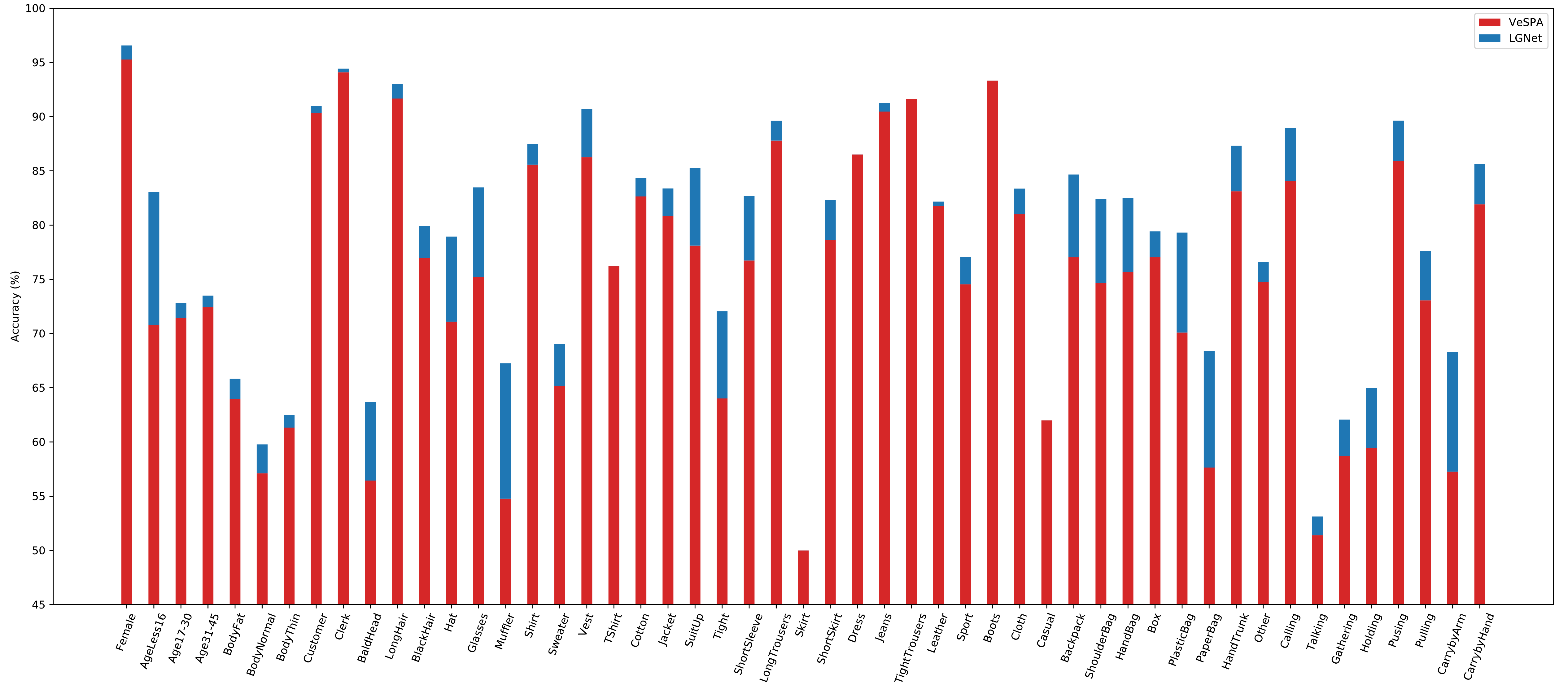}
\end{center}
\caption{Mean accuracy comparison between our LG-Net and VeSPA on 51 
attributes of RAP dataset. LG-Net significantly surpasses VeSPA in most 
attributes, especially attributes related to associated objects. The 
improvement in mean accuracy for object-type attributes proves the 
effectiveness of our designed localization guided architecture.}
\label{fig:fig3}
\end{figure}
To highlight the performance of the attribute-specific localization 
result of the proposed method, we compare the mean accuracy of 
51 attributes in RAP dataset between our LG-Net and VeSPA, the 
previous state-of-the-art, as shown in Figure 3. The result shows 
that our method performs 
better than VeSPA in most attributes, especially attached objects, for 
example, "backpacks", "plastic bags" and "calling". 
The attached objects are closely related to our proposed localization 
guided method since attached objects are more sensitive to the 
relative positions and viewpoints. Moreover, our method performs 
slightly better in action recognition including "talking" and "gathering" because 
the LG-Net utilizes both global and local information for attribute 
predictions. 

\section{Ablation Study}
To better demonstrate the effectiveness and advantage of our 
designed architecture, we apply component-wise ablation studies 
to explicitly address the contribution of each block.

\textbf{Mapping Resolution.} The original Inception-v2 produce 
feature maps in a mapping resolution of $7\times7$. We stretch 
the feature maps into $14\times14$ by changing the stride in 
\textit{inception-4e}. The replacement leads to minor computational 
inefficiency, but the localization performance is apparently 
improved. Better localization result is helpful for our future 
studies on local feature extractions.

Table 2 shows the comparison between the result of the model with 
$14\times14$ feature map and the $7\times7$ baseline architecture. 
The result shows that network with $14\times14$ feature map surpasses 
$7\times7$ result by 0.92\% in mA and 0.45\% in Accuracy.

\textbf{Dilated Convolution.} Dilated convolution~\cite{yu2016weakly} 
is widely adopted in some recent-proposed networks for person 
re-identification~\cite{li2017learning} and some other tasks 
~\cite{gao2017im2flow} 
to enlarge the receptive field. Dilated convolution is adopted 
in the \textit{inception-4e} to maintain $14\times14$ feature maps 
instead of directly replacing convolution stride by one. We compare 
the dilated convolution version with the ordinary convolution as shown 
in table 2. We compare dilated convolutions with dilation ratio 1, 2, 
and change dilation ratio to 2 on one branch only. We adopt the version 
with dilation ratio 1 and 2 for ultimate performance.

\textbf{Affinity Strategy.} For the fusion network, one key factor 
that affects the result is the proposal weighting strategy. We 
tried two strategies to calculate the 
affinity between bounding box extracted from Class Activation Map 
and Region of Interests. The adopted version is to calculate the 
Intersection over Union (IoU) as the weight of proposals. Apart from 
this version, we also tried to weight the proposal by directly counting 
the overlapping area of the corresponding bounding boxes. Table 2 
demonstrates the effectiveness of using IoU as the weight. IoU brings 
0.84\% and 0.42\% boost in mA and Accuracy respectively.

\begin{table}[]
\centering \small
\begin{tabular}{|l|ccccc|}
\hline
Dataset for Ablation Study  & \multicolumn{5}{c|}{RAP} \\ \hline
Method                   & mA & Accu & Prec & Recall & F1 \\ \hline
$7\times7$ Feature Map   & 75.43 & 65.94 & \textbf{79.78} & 77.05 & 78.39 \\ \cline{1-1}  
$14\times14$ Feature Map & \textbf{76.35} & \textbf{66.39} & 78.53 & \textbf{78.98} & \textbf{78.75} \\ \hline\hline
Dilation ratio = 1   & 76.35 & 66.39 & 78.53 & 78.98 & 78.75 \\ \cline{1-1}  
Dilation ratio = 2   & \textbf{77.05} & 66.49 & 78.39 & \textbf{79.33} & 78.86 \\ \cline{1-1}  
Dilation ratio = 1 and 2 & 76.79 & \textbf{67.32} & \textbf{79.69} & 79.28 & \textbf{79.49} \\ \hline\hline
Overlapping Area as Proposal Weight   & 77.84 & 67.58 & 80.22 & 79.32 & 79.77 \\ \cline{1-1}  
IoU as Proposal Weight    & \textbf{78.68} & \textbf{68.00} & \textbf{80.36} & \textbf{79.82} & \textbf{80.09} \\ \hline\hline
LG-Net without localization   & 77.88 & 63.59 & 75.21 & 78.63 & 76.88 \\ \cline{1-1}  
LG-Net    & \textbf{78.68} & \textbf{68.00} & \textbf{80.36} & \textbf{79.82} & \textbf{80.09} \\ \hline
\end{tabular}
\caption{Ablation Study: Effects of Components.}
\end{table}

\textbf{Localization Guidance.} Localization Guidance Module is the 
core of our network. To better illustrate the effectiveness of weight 
proposals by the affinity between proposals and localization boxes, we 
modify our network by removing the localization generator as well as 
the localization guidance module. All proposals are weighted evenly in 
the modified network. Table 2 shows the comparison between the 
modified LG-Net without localization guidance and the original LG-Net. 
Result suggests that localization module significantly improves the 
accuracy by 4.4\%, which means that accurate localization of attributes 
is vital to the recognition of attributes for local feature extraction.

\section{Conclusion}
In this paper, we present a novel Localization Guided Network for 
the local feature extraction and utilization with attribute-specific 
localization guidance. Our LG-Net indicate that spatial information 
of attributes is helpful to the utilities of local features. Further 
ablation studies show the effectiveness of each component of our 
localization network and indicate that localization guidance is the key 
factor of success. Our network outputs the localization result parallel 
to the attribute predictions, and achieves the state-of-the-art results 
on the two largest pedestrian attribute datasets. 

\bibliography{egbib}
\end{document}